\documentclass[fleqn,11pt]{wlscirep}
\setcitestyle{sort&compress,open={},close={},citesep={,},numbers,super}
\usepackage[utf8]{inputenc}
\usepackage[T1]{fontenc}
\usepackage{bm}
\usepackage{algorithm} 
\usepackage{algpseudocode} 
\usepackage{graphicx}
\usepackage{float} 
\usepackage{subfigure}
\usepackage{xspace}
\usepackage{amsmath,amsfonts,amssymb}
\usepackage{pdfpages}
\usepackage{tabularx}
\usepackage{makecell}
\usepackage{booktabs}
\usepackage{threeparttable}
\usepackage{microtype}
\usepackage[misc]{ifsym}

\usepackage{amsmath}
\usepackage{array}

\usepackage{booktabs}
\usepackage{multirow}
\usepackage{amsfonts}
\usepackage{amssymb}

\usepackage{nicefrac}
\usepackage{url}
\usepackage[switch]{lineno}

\usepackage{setspace}
\usepackage{enumitem}

\makeatletter
\newcommand{\ssymbol}[1]{$^{\@fnsymbol{#1}}$}
\makeatother

\usepackage{soul}

\title{Large Language Models Leverage External Knowledge to Extend Clinical Insight Beyond Language Boundaries}

\author[1]{Jiageng Wu, MS}
\author[2 \Letter]{Xian Wu, PhD}
\author[2]{Zhaopeng Qiu, MS}
\author[1]{Minghui Li, BM}
\author[2]{Yingying Zhang, PhD}
\author[2]{Yefeng Zheng, PhD}
\author[1,3 \Letter]{Changzheng Yuan, PhD}
\author[4 \Letter]{Jie Yang, PhD}

\affil[1]{School of Public Health, Zhejiang University School of Medicine, Hangzhou, China;}
\affil[2]{Jarvis Research Center, Tencent YouTu Lab, China}
\affil[3]{Harvard T.H. Chan School of Public Health, Boston, USA,}
\affil[4]{Harvard Medical School, Boston, MA}
\begin{abstract}
\textbf{Objectives:}
Large Language Models (LLMs) such as ChatGPT and Med-PaLM have excelled in various medical question-answering tasks. However, these English-centric models encounter challenges in non-English clinical settings, primarily due to limited clinical knowledge in respective languages, a consequence of imbalanced training corpora. We systematically evaluate LLMs in the Chinese medical context and develop a novel in-context learning framework to enhance their performance. 

\textbf{Materials and Methods:}
The latest China National Medical Licensing Examination (CNMLE-2022) served as the benchmark. 
We collected 53 medical books and 381,149 medical questions to construct the medical knowledge base and question bank. The proposed Knowledge and Few-shot Enhancement In-context Learning (KFE) framework leverages the in-context learning ability of LLMs to integrate diverse external clinical knowledge sources. We evaluated KFE with ChatGPT(GPT3.5), GPT4, Baichuan2-7b, and Baichuan2-13B in CNMLE-2022 and further investigated the effectiveness of different pathways for incorporating LLMs with medical knowledge from seven distinct perspectives. 

\textbf{Results:}
Directly applying ChatGPT failed to qualify for the CNMLE-2022 at a score of 51. 
Cooperated with the KFE framework, the LLMs with varying sizes yielded consistent and significant improvements. 
The ChatGPT's performance surged to 70.04 and GPT-4 achieved the highest score of 82.59. This surpasses the qualification threshold (60) and exceeds the average human score of 68.70, affirming the effectiveness and robustness of the framework. It also enabled a smaller Baichuan2-13B to pass the examination, showcasing the great potential in low-resource settings. 

\textbf{Discussion and Conclusion:}
This study shed light on the optimal practices to enhance the capabilities of LLMs in non-English medical scenarios. By synergizing medical knowledge through in-context learning, LLMs can extend clinical insight beyond language barriers in healthcare, significantly reducing language-related disparities of LLM applications and ensuring global benefit in this field. 


\end{abstract}

\begin{document}
\flushbottom
\maketitle
\doublespacing

\thispagestyle{empty}

\section*{INTRODUCTION}
\label{sec:intro}
Large Language Models (LLMs),\cite{llm-survey-2023} especially the Generative Pre-trained Transformer (GPT),\cite{llm-gpt2-2019} have presented the substantial potential in healthcare applications to enhance medical research and improve clinical services.\cite{llm-med-NatMed} The extensive pre-training on large-scale corpora equips LLMs with strong capabilities for text comprehension and generation,\cite{bert-2018,llm-text-generation-acl2019} and also encodes a broad spectrum of knowledge \cite{llm-knowledge-acl2019} to support reasonable medical analyses.\cite{chatgpt-medical-reasoing-jno}
Moreover, their inferential capabilities are further augmented by comprehensive supervised fine-tuning across diverse tasks,\cite{instruction-tuning-2022} thereby bolstering clinical diagnosis and decision-making.\cite{llm-engine-nature}
Benefiting from the text-to-text architecture of GPT, which allows for direct interaction via textual prompts,\cite{llm-gpt-unilm} LLMs offer flexibility in constructing question-and-answer systems with prompts. This feature significantly enhances usability for individuals lacking coding experience,\cite{chatgpt-resource-lancet-2023} particularly in leveraging advanced AI techniques in digital medicine, potentially revolutionizing the whole field.\cite{chagpt-review-NEJM-2023} 
For physicians, GPTs can provide assistance in disease diagnosis,\cite{chatgpt-eye-diagnosis} medication recommendation,\cite{chatgpt-drug-jamia} drafting summary,\cite{chatgpt-summary-2023} and instruction generation.\cite{chatgpt-consulation-2023} This can relieve the heavy workload of doctors and alert the misdiagnoses and underdiagnoses. 
For patients, especially those in areas with less accessible medical resources,\cite{chatgpt-lmic-lancet} GPTs can serve as versatile medical assistants, capable of analyzing real-world medical problems and providing general suggestions in a user-friendly and timely manner.\cite{medical-cardio-jama, suggestion-ph-2023}
Encouragingly, recent studies proved that several advanced LLMs can attain the level of proficiency in medical knowledge as a junior general practitioner, which is evidenced by qualifying for the United States Medical Licensing Examination (USMLE) with high scores. \cite{llm-gpt4medical-2023, medical-chatgpt-usmle, medical-palm-usmle} 

\renewcommand\floatpagefraction{.9}
\renewcommand\topfraction{.9}
\renewcommand\bottomfraction{.9}
\renewcommand\textfraction{.1}
\begin{figure*}[!tp]
    \centering
    \includegraphics[width=16.6cm]{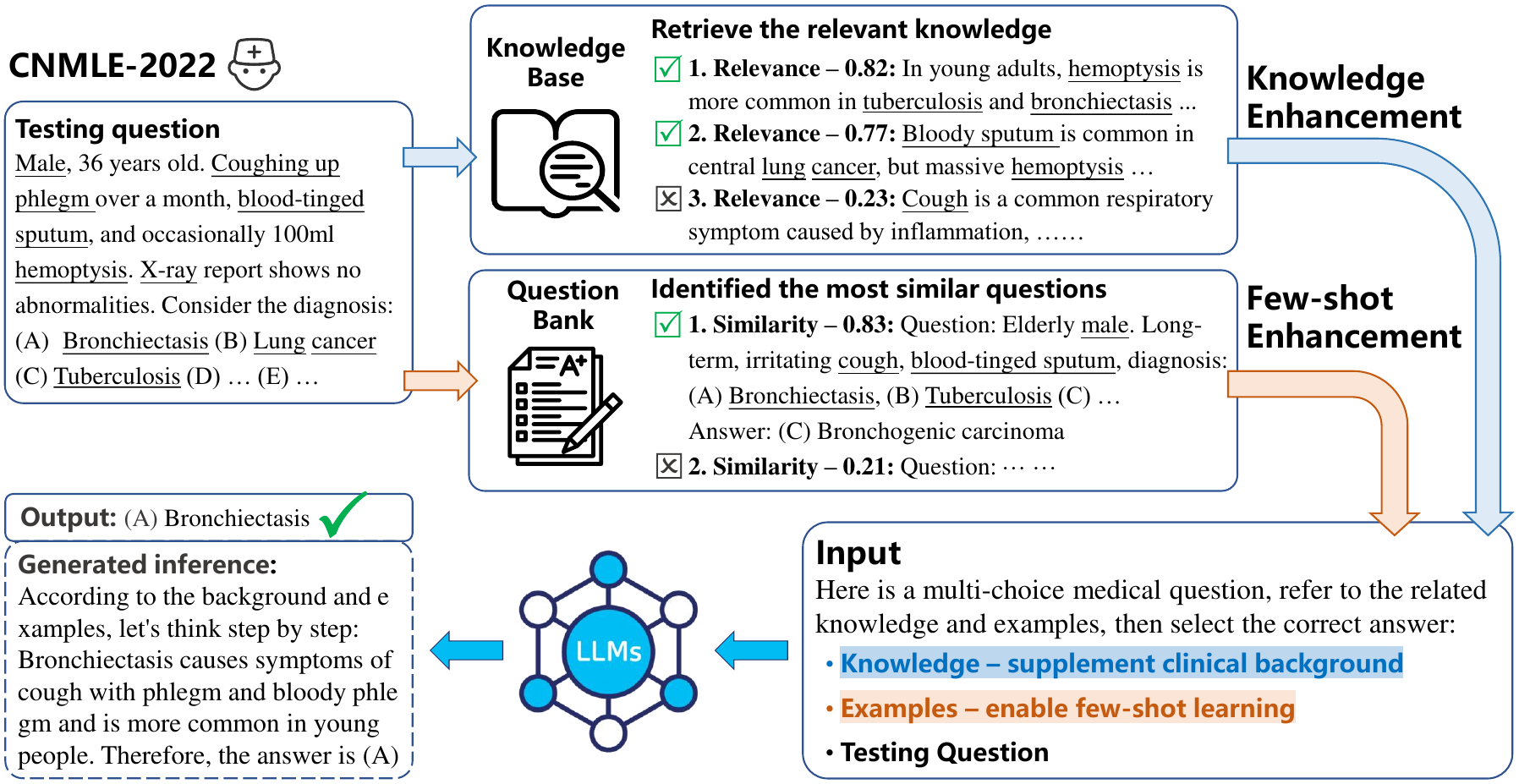}
    \vspace{-0.5pt}
    \caption{\textbf{Workflow of knowledge and few-shot enhancement framework (KFE)}. Given the question stem and five candidate options, KFE retrieves the relevant knowledge from a comprehensive knowledge base and identifies similar questions from a historical question bank. The integrated prompt consists of specific instructions, pertinent knowledge, exemplar cases, and a testing question. Benefiting from the LLM's in-context learning capabilities, KFE extends the clinical insight of LLM in non-English settings and bridges the language barriers in healthcare applications. Finally, LLM generates a predicted answer. }
    \label{fig: Overall}
    \vspace{-0.36cm}
\end{figure*}

Despite significant advancements in LLMs, their development and validation predominantly rely on English-centric, general-domain datasets.\cite{llm-nonenglish-book, llm-evaluation-multilingual-safety} However, there are over 7,000 languages worldwide beyond English, encompassing 81.35\% of the global population,\cite{MostSpokenLanguages} many of whom may find it challenging to benefit from these advanced models.\cite{llm-evaluation-multilingual-multitask} Given that approximately 90\% of data for training LLMs is in English \cite{llm-dataset-non-english} and non-English medical corpora are even more scarce,\cite{llm-domain-specific-pt} this language imbalance of training corpora poses challenges for the extensions of LLMs in non-English clinical scenarios.\cite{chatgpt-resource-lancet-2023} The insufficiency of data and training impedes LLMs in understanding the intricate concepts \cite{medical-reason} and encoding the necessary clinical expertise in non-English environment,\cite{llm-medical-jama} which is essential for accurate clinical inference. 

While existing studies tried to integrate clinical knowledge into LLMs during the training stage, as exemplified by Med-PaLM \cite{medical-palm-usmle} and GatorTronGPT,\cite{llm-medical-GatorTronGPT} these primarily English-centric LLMs lack the generalizability needed for diverse linguistic contexts, thus limiting their global applicability in healthcare.\cite{chatgpt-resource-lmic} Moreover, they necessitate extensive, high-quality datasets and substantial computational resources, rendering them impractical for low-resource groups or countries.\cite{chatgpt-resource-lancet-2023} 
How to effectively integrate various types of healthcare knowledge into LLMs across different language boundaries remains under-investigated. Existing studies have revealed a marked performance decline of LLMs in non-English medical scenarios \cite{chatgpt-exam-china, chatgpt-exam-japan} and raised concern about the potential bias brought by the under-represented data.\cite{chatgpt-eye-diagnosis, chatgpt-exam-japan-bias} Nonetheless, these studies primarily focused on the direct application and performance evaluation, without delving into how to enhance the capability of LLMs in non-English medical contexts. 
Consequently, it remains unclear 1) what the best practice for using GPTs in non-English medical scenarios is, 2) how well these models perform in this scenario, and 3) how to incorporate medical knowledge to bolster GPTs. 

To address the above challenges, we apply  GPTs to the China National Medical Licensing Examination (CNMLE) \cite{cnmle-intro-web} and investigate effective approaches to improve performance by integrating medical knowledge. 
Similar to USMLE, CNMLE is an essential qualifying examination to become a certified doctor in China, covering knowledge from 20 medical subjects in four areas: clinical medicine, preclinical medicine, medical humanities, and preventive medicine.\cite{cnmle-outline-web} Passing CNMLE requires not only a deep and broad understanding of medical knowledge but also the ability to analyze Chinese real-world clinical cases.\cite{CNMLE-med3r-2018} In this work, we propose \textbf{K}nowledge and \textbf{F}ew-shot \textbf{E}nhancement In-context Learning (KFE) framework to leverage the in-context learning ability of LLMs \cite{llm-gpt3-2020} with the domain-specific knowledge. The KFE framework includes two in-context learning strategies: 1) Knowledge Enhancement: we build a medical knowledge base as a source to provide background knowledge; 2) Few-shot Enhancement: we collect a dataset of historical questions and answers of CNMLE as a question bank to provide few-shot examples. Four types of Chain-of-Thought (CoT) strategies \cite{cot-first-wei2022, cot-step-kojima2022} are designed and examined to enrich the information of retrieved questions. 
Experiment results demonstrate that LLMs with KFE significantly improve their performance to qualify for CNMLE with high scores, even than human candidates. The GPT4 with KFE yielded the highest score of 82.59. Furthermore, extensive ablation studies were conducted to systematically investigate the effectiveness of different pathways in incorporating LLMs with various medical knowledge.

\section*{MATERIALS AND METHODS}
\subsection*{Dataset}
The CNMLE serves as the official qualification examination for clinicians in China, with over half a million medical practitioners participating annually.\cite{cnmle-intro-web} CNMLE evaluates not only the proficiency of medical knowledge but also the practical skills in real clinics. Candidates must complete five years of medical education and additionally undergo a one-year clinical practice assessment. 
The CNMLE encompasses various types of questions from multiple medical disciplines.\cite{cnmle-outline-web} Most of them can be reformulated as objective multiple-choice questions, each presenting five candidate options. 
The passing score for the CNMLE is set at 60, indicating that a candidate must answer at least 60\% of the questions correctly to qualify. Considering the skills assessed, these questions predominantly be grouped into two categories: medical knowledge questions (MK) and case analysis questions (CA). The MK questions demand a comprehensive grasp of medical concepts and terminology. Meanwhile, the CA questions present real-world clinical cases that necessitate accurate diagnosis or treatment based on clinical factors like patient symptoms, examination results, medical history, etc., emphasizing applying medical knowledge in clinical practice. 

To ensure that the testing questions were not previously included in the training set for the LLMs, we curate 494 multi-choice questions from the most recent CNMLE held in Aug 2022 for evaluation. Given that the datasets used for retrieval were collected before Sep 30th, 2021, there is no risk of label leakage. 

\subsection*{Problem Formulation}
\label{sec:problem}
Each medical question is represented in the form of a triple $\{Q, O, A\}$ where $Q$ refers to the question stem, $O=\{o_1, ..., o_5\}$ refers to the five candidate options, and $A$ refers to the answer which is one option in $O$. Therefore, in the context of the GPTs, answering medical problems can be formulated as estimating the probability of generating the correct answer $P(A|Q, O)$ given question $Q$ and options $O$. 

To correctly guide the GPT in solving the medical question, specific instructions $I$ are provided to describe the task. We use two types of instructions here:
\begin{itemize}[itemsep=1pt,topsep=0pt,parsep=0pt]
    \item \textbf{Direct Instruction}: \textit{``Here is a multi-choice question about medical knowledge, please output the correct answer according to the question.''} We refer to this direct instruction as $I_{direct}$, which only requires the GPT to generate the correct answer. Then, the task can be formulated as estimating the probability $P(A|Q, O, I_{direct})$. 
    \item \textbf{Instruction with inference}: \textit{``Here is a multi-choice question about medical knowledge, please analyze it in a step-by-step fashion and deduce the correct answer.''} We refer this kind of instruction to $I_{steps}$, which requires the GPT to generate both the correct answer as well as the detailed inference steps. Then, the task can be formulated as estimating the probability $P(A|Q, O, I_{steps})$. This kind of instruction is motivated by CoT, which has been found effective in generating the correct answer \cite{cot-first-wei2022}. 
\end{itemize}

\subsection*{Overall Framework}
To further improve the performance, we propose the \textbf{K}nowledge and \textbf{F}ew-shot \textbf{E}nhanced In-context Learning (KFE) framework. Figure \ref{fig: Overall} illustrates the KFE framework, comprising two primary modules: {\em Medical Knowledge Retriever} and {\em Question Bank Retriever}. Given a question stem and options, the Medical Knowledge Retriever extracts the relevant medical knowledge from an established medical knowledge base, which is then incorporated into the prompts as background knowledge. In parallel, the Question Bank Retriever acquires similar questions and their corresponding answers from a curated Question Bank, which are integrated into the prompts to enable few-shot learning. 
Accompanying explicit instruction and bolstered by pertinent knowledge and questions, the GPT leverages its in-context learning ability to extend clinical insight from the background and learn the capability of problem-solving as exemplified in the sample questions, thereby generating answers to medical queries correctly. 
\subsection*{Knowledge Enhancement}
\label{sec:knowledge}
We construct a comprehensive medical knowledge base that is generated from 53 textbooks of People's Medical Publishing House. These books are recommended textbooks for most medical schools in China and their quality is well assured. We split the content of each book into text pieces by leveraging the structure of the books. In total, we acquired 68,962 pieces of text, with an average length of 130 tokens per text. 

To infer the correct answer to a question, both the questions and all candidate options contain critical information. In many cases, it is required to combine the question and the candidate option together to form a completed medical context. Therefore, we concatenate each option $o_i \in O$ with its corresponding question $Q$, which serves as a query, to retrieve the most relevant pieces of knowledge $k_i$ from the knowledge base: 
$$
k_i=\mathop{\arg\max} R_K(k | (q \mathbin\Vert a_i)), 
$$
where $q \mathbin\Vert a_i$ refers to the concatenation of the question with one option, $R_K$ represents the knowledge retrieval engine that returns the most relevant knowledge $k_i$ given $(q \mathbin\Vert a_i)$. To enhance the efficiency of retrieval, we employ BM25, which is an extension of TF-IDF, as our retrieval engine. BM25 has been proven effective in retrieving examples for in-context learning in QA tasks, sometimes even better than BERT-based methods.\cite{retrieve-bm25-better}

As a result, for all five pairs of question-and-option, we can collect 5 pieces of knowledge $k = \{k_1,\dots, k_5\}$.
This strategy ensures that the retrieved knowledge is relevant to the context of the question and provides more concentrated and useful background knowledge.

\subsection*{Few-shot Enhancement}
\label{sec:fewshot}
We initially curate a sizable medical question bank, denoted as $B = \{b_1, b_2, \dots, b_m\}$, encompassing a significant volume of medical questions derived from historical CNMLE, textbooks, and reference materials. In total, we build a medical question bank with 381,149 questions. Each instance in this question bank includes a question stem, five candidate options, and the correct answer. 

Similar to the knowledge retrieval approach, we also query similar examples from the question bank by combining the question and options together. However, instead of enumerating all question and option pairs, we concatenate the question with all options to match similar problems in the question bank. Specifically, we concatenate the question with all options to generate the context $(q \mathbin\Vert O)$, which is used to search for the top-$k$ similar examples from question bank by BM25: 
$$
b_{q}=\mathrm{\arg\max}_1^{k}R_B(b | (q \mathbin\Vert O)), 
$$ 
where $k$ is the number of examples and $R_B$, denotes the retrieval engine that returns relevant examples. 

\begin{figure}[t]
    \centering
    \includegraphics[width=14cm]{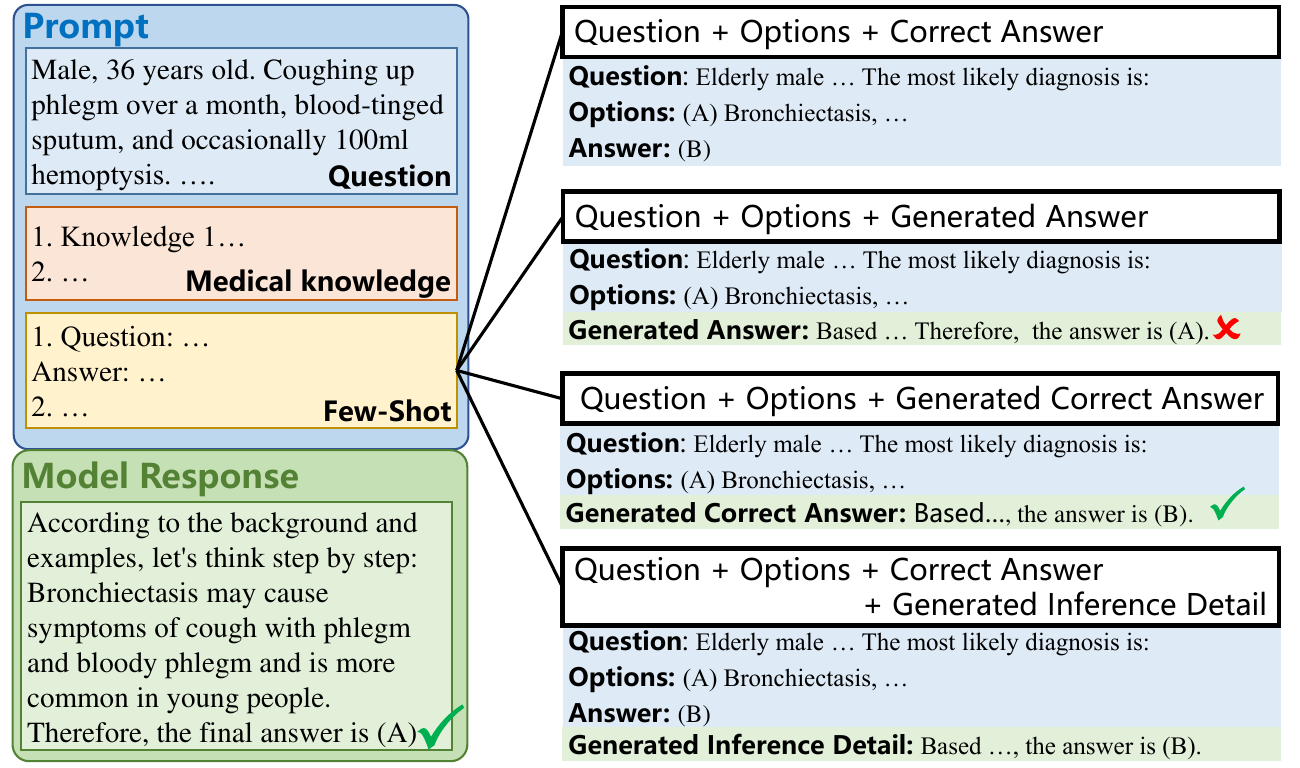}
    \vspace{-0.5pt}
    \caption{Four strategies of few-shot enhancement.}
    \label{fig: few-shot}
    \vspace{-0.3cm}
\end{figure}

After retrieving relevant examples, we can leverage the few-shot strategy to enhance the problem-solving capabilities of LLMs. As shown in Figure \ref{fig: few-shot}, we propose four strategies to add few-shot enhancement:
\begin{itemize}[itemsep=1pt,topsep=0pt,parsep=0pt]
    \item {\bf Question + Options + Correct Answer:} for each retrieved example, we concatenate the question $Q$, all candidate options $O$, and the correct answer $A$ together that is used as the few-shot part in the prompt.
    \item {\bf Question + Options + Generated Answer:} for each retrieved example, we first send the question $Q$ and all options $O$ to the GPT to generate the answer. The acquired answer is then appended back to the question and options as the few-shot part of the prompt. This approach brings additional computational cost, but the correct answer is no longer required. Furthermore, despite the generated incorrect answer, it can still guide the GPT to learning the format and solution of medical questions, which may barely hurt the performance \cite{cot-false-2022}.
    \item {\bf Question + Options + Generated Correct Answer:} Different from the above strategy, we only keep the examples with correctly generated answers. For those questions with incorrectly generated answers, we remove them and pick other examples with lower relevance from the Question Bank.
    \item {\bf Question + Options + Correct Answer + Generated Inference Detail:} In this case, we send the triplet $\{Q, O, A\}$ to the GPT and instruct it to generate the inference details of why the correct answer $A$ is chosen. The generated inference details are concatenated with questions, options, and the correct answer to form the few-shot section in the prompt.
\end{itemize} 

\subsection*{Model settings}
We chose GPT3.5-Turbo, the driving engine behind the widely used ChatGPT, as the primary LLM to evaluate. It includes 175B parameters and demonstrated exceptional performance across diverse tasks.\cite{llm-nlp-diyi2023} Moreover, to verify the robustness and efficacy of our framework, we also evaluated the larger LLM (GPT-4) and the smaller LLM (Baichuan2-7B/13B). 
GPT-4, which is estimated to have 1.7 trillion parameters and stands as the current largest and most powerful model.\cite{llm-gpt4medical-2023} Baichuan2(BC2), an open-source LLM, consists of either 7 billion or 13 billion parameters \cite{baichuan2}, which can be conveniently deployed on common PCs. It has been trained from scratch on 2.6 trillion multilingual tokens, including English, Chinese, Spanish, Arabic, and others. Owing to the training in the extensive Chinese data, Baichuan2 exhibits superior performance across various Chinese tasks. 

GPT3.5 and GPT4 were conducted by calling OpenAI’s official API. Unless specified, all experiments used the same parameters and were tested with the same model version (See Supplementary Table 1). We set the inference temperature to 0 to make the response deterministic. All other parameters are set to default (Supplementary Table 2). 
\subsection*{Experiment settings}
To fully reveal the performance of LLMs, we evaluate several competitive baselines as well as different variants of the proposed KFE model as follows:

\begin{itemize}[itemsep=1pt,topsep=0pt,parsep=0pt]
\item \textbf{Fully Supervised Model:} SeaReader \cite{CNMLE-seareader-2018} and Med3R \cite{CNMLE-med3r-2018} are fully supervised deep learning models and trained on 230k and 270k medical questions respectively. They formulate medical questions as reading comprehension tasks that extract information from five relevant documents to determine the answer. Med3R underwent additional pre-training on many medical books, achieving state-of-the-art results (SOTA) on CNMLE-2017. 

\item \textbf{GPT with Direct Instruction:} Here we use the direct instruction $I_{direct}$.
To investigate the effect of different components of KFE, we conducted extensive experiments on various strategies: {\em Zero-shot} denotes the basic approach without any enhancement; {\em Few-shot} denotes the approach with only few-shot enhancement; {\em Knowledge Enhancement} denotes the approach with only knowledge enhancement and KFE denotes the complete proposed framework. supplementary Figure 1 presents the integrated prompt for different methods. 

\item \textbf{GPT with Instruction with Inference Steps:} Here we use the instructions with inference steps $I_{steps}$. The rest settings are the same as {\em GPT with Direct Instruction}. Here we aim to investigate whether the generated inference details can enhance problem-solving ability.

\item \textbf{Human:} Beyond the qualified score of 60, we also compared the average score of medical students from a China top-tier hospital who participated in the CNMLE-2022 \cite{chatgpt-exam-china}, serving as a strong human benchmark. 

\end{itemize}

\section*{RESULTS}
\subsection*{Model performance in CNMLE} 
We compare the performance of different methods in CNMLE in Table \ref{table: Performance all}. Supplementary Figure 1 presents the integrated prompt for different methods. The fully supervised approaches outperform the GPT-based approaches, as these supervised approaches are specially tailored for medical exams and cannot be applied to other medical tasks. In addition, these supervised models are trained with more than 230k historical questions. The training process is notably time-consuming. While the GPT-based approaches require less than 10 few-shot examples and do not need to fine-tune the backbone GPTs.

Among GPT-based approaches, the larger LLMs (GPT3.5 and GPT4) with proposed KFE not only passed the latest CNMLE (70.04 and 82.59) but also surpassed the average score of medical students (68.70). Moreover, the smaller BC2-13B with KFE also qualified for the CNMLE-2022. These consistent and significant improvements across LLMs with varying sizes further validate the superiority of KFE, underscoring its applicability in limited-resources settings. Supplementary Materials 1 and 2 illustrate case studies of knowledge and few-shot enhancement.

Our result reveals that both the knowledge and few-shot enhancement substantially improve the overall performance. The integration of either enhancement markedly elevates the capabilities of basic GPTs. Another notable finding is that the GPT with $I_{direct}$ outperforms GPT with $I_{steps}$, this discrepancy may stem from the mistakes and hallucinations present in generated inferences, which potentially misguide the GPTs towards incorrect conclusions. 



\begin{table}[htbp]\centering
  \caption{Performance of different methods in CNMLE.}
  \scriptsize
  \resizebox{0.69\linewidth}{!}{
  \begin{tabular}{lccc}
    \toprule
    \textbf{Method} &\makecell[c]{\textbf{Acc-MK}(\%)} &\makecell[c]{\textbf{Acc-CA}(\%)} & \makecell[c]{\textbf{Acc-All}(\%)} \\
    \midrule
    \makecell[l]{\textbf{Fully Supervised Model}} &  &  &  \\
    SeaReader \cite{CNMLE-seareader-2018} & - & - & 57.8 \\
    Med3R (Domain pre-training)\cite{CNMLE-med3r-2018} & \textbf{77.34} & \textbf{75.00} & \textbf{76.00} \\
    \midrule
    \makecell[l]{\textbf{GPT3.5 with Instruction $I_{direct}$}} &  &  &    \\
    Zero-shot & 49.17	& 52.08	& 51.01 \\
    Few-shot & 65.75 & 62.30 & 63.56 \\
    Knowledge Enhancement & 68.51 & 58.15 & 61.94 \\
    KFE & \textbf{72.93} & \textbf{68.37} & \textbf{70.04} \\
    \midrule
    \makecell[l]{\textbf{GPT3.5 with Instruction $I_{steps}$}} &  &  &    \\
    Zero-shot & 51.93 & 52.08 & 52.02 \\
    Few-shot & 59.12 & 56.87 & 57.69 \\
    Knowledge Enhancement & \textbf{72.38} & 54.95 & 61.34 \\
    KFE & 66.30 & \textbf{64.86} & \textbf{65.38} \\
    \midrule
    \makecell[l]{\textbf{GPT4 with Instruction $I_{direct}$}} &  &  &    \\
    Zero-shot & 66.30	& 70.93	& 69.23 \\
    Few-shot & 81.77 & \textbf{81.47} & 81.58 \\
    Knowledge Enhancement & 72.93 & 70.93 & 71.66 \\
    KFE & \textbf{84.53} & \textbf{81.47} & \textbf{82.59} \\
    \midrule
    \makecell[l]{\textbf{BC2-7B with Instruction $I_{direct}$}} &  &  &    \\
    Zero-shot & 49.17 & 49.52 & 49.39 \\
    Few-shot & 48.07 & 54.95 & 52.43  \\
    Knowledge Enhancement & \textbf{56.91} & 45.69 & 49.80  \\
    KFE & 55.80 & \textbf{58.15} & \textbf{57.29}  \\
    \midrule
    \makecell[l]{\textbf{BC2-13B with Instruction $I_{direct}$}} &  &  &    \\
    Zero-shot & 52.49 & 57.83 & 55.87 \\
    Few-shot & 50.28 & 61.66 & 57.49  \\
    Knowledge Enhancement & \textbf{61.88} & 56.55 & 58.50  \\
    KFE & 59.67 & \textbf{61.98} & \textbf{61.33}  \\
    \midrule
    \makecell[l]{\textbf{Human}} &  &  &    \\
    Passing core & -  & -  & 60.00   \\
    Average of medical students\cite{chatgpt-exam-china} & - & - & \textbf{68.70} \\
    \bottomrule
    \end{tabular}
    \label{table: Performance all}
    }
    \vspace{-0.3cm}
\end{table}

\subsection*{Ablation Studies and Analysis}
In this section, we conduct ablation studies and analysis from the following perspectives: 1) we evaluate different strategies for few-shot enhancement; 2) we evaluate the contribution of generated inference details with different lengths; 3) we study the contribution of different numbers of few-shot examples; 4) we compare the performance of different prompting strategies $I_{direct}$ and $I_{steps}$; 5) the effectiveness of {\em Medical Knowledge Base}; 6) the effectiveness of {\em Question Bank Retrieval}; 7) limitations on length and characters of the model responses. 

\subsubsection*{Effect of Different Strategies for Few-shot Enhancement}
\begin{figure*}[htbp]
    \centering
    \includegraphics[width=0.96\linewidth]{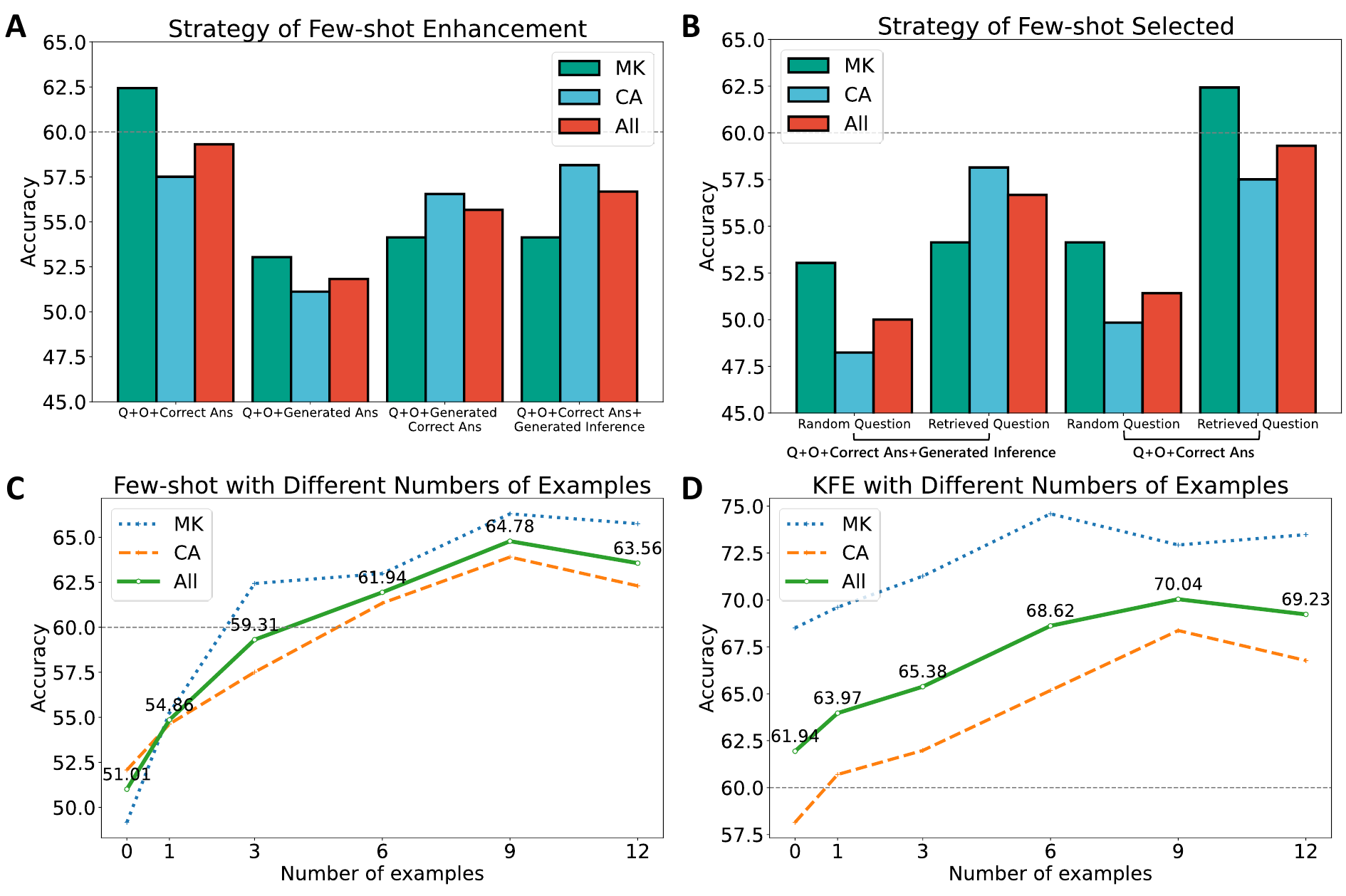}
    \caption{\textbf{Ablation studies and analysis.} \textbf{A} The strategy of few-shot Enhancement. \textbf{B} The strategy of few-shot selected. \textbf{C} Few-shot enhancement with different numbers of examples. \textbf{D} KFE with different numbers of examples. \textbf{MK} and \textbf{CA} refer to questions about medical knowledge and clinical case analysis, respectively. }
    \label{fig: ablation studies}
    \vspace{-0.3cm}
\end{figure*}

Figure \ref{fig: few-shot} displays four different strategies for few-shot enhancement and Supplementary Material 3 demonstrates a case study of these strategies. 
As shown in Figure \ref{fig: ablation studies}(A), the \textit{Q+O+Correct Ans} achieved the highest score of 59.31. Compared to the other three strategies, \textit{Q+O+Correct Ans} uses the least generated information from GPT in composing the prompts. 
Another observation is that \textit{Q+O+Generated Ans} (51.82) underperformed \textit{Q+O+Generated Correct Ans} (55.67) by a large margin. These two observations showed that the presence of generated content may impair performance and even lead to a result worse than Zero-shot (52.02), which is consistent with previous in-context learning approaches \cite{cot-right-2022,cot-auto-zhang2022} and in conflict with.\cite{cot-false-2022} This may be due to that the generated content contains mistakes and answering questions in CNMLE requires high precision. Therefore integrating these unconfirmed auto-generated contents in prompts could mislead the GPTs and in turn generate incorrect answers. 

\subsubsection*{Analysis of Generated Inference Details with Varied Length}
\begin{figure}[htbp]
    \centering
    \includegraphics[width=9.9cm]{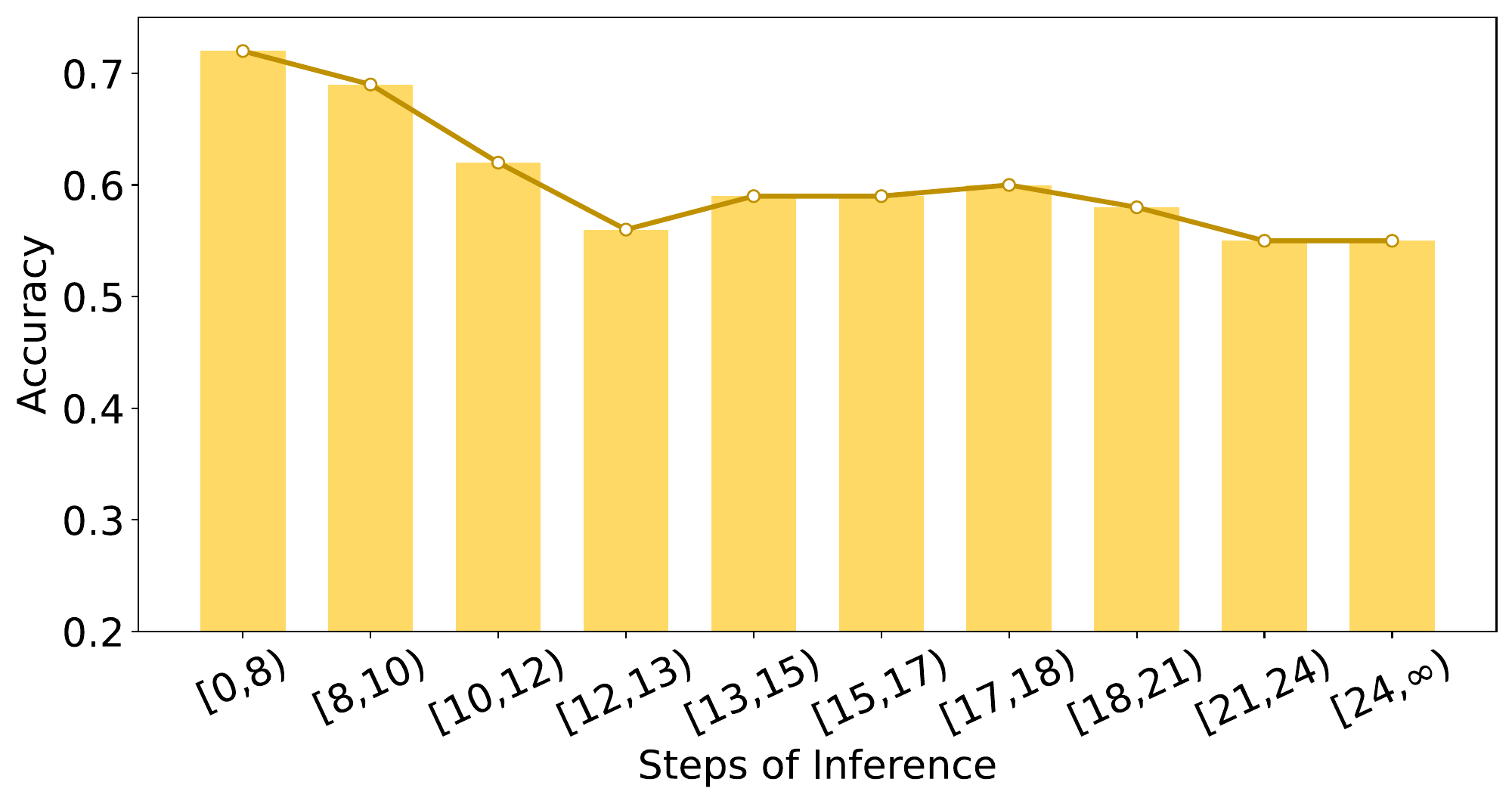}
    \caption{Performance w.r.t. varied length of generated inference details. } 
    \label{fig: steps}
    \vspace{-0.66cm}
\end{figure}

Given the generated inference, we use the metric {\em Inference Step} to measure its complexity as introduced in .\cite{cot-complex-2022} Specifically, we first conduct sentence segmentation on generated inference details and allocate them into ten buckets according to the number of sentences. As shown in Figure \ref{fig: steps}, the shorter inference steps yield better accuracy on medical examination which is different from the findings in \cite{cot-complex-2022}, which reported that GPT achieves substantially better performance on reasoning tasks with more inference steps. This may be due to that longer inference steps contain more mistakes and hallucinations.

\subsubsection*{Effect of Different Numbers of Few-shot Examples} 
We investigate how the performance varies with an increase in the number of few-shot examples. Here we choose the optimal \textit{Q+O+Correct Ans} strategy for few-shot enhancement. Notably, due to the limitation of the maximum token of the GPT model (4096 tokens maximal), we have increased the number of examples as much as possible and the maximal examples in Few-shot and KFE are both 12. 
As shown in Figure \ref{fig: ablation studies}(C) and (D), a significant performance improvement is observed with the increase in example counts. Specifically, the Few-shot method demonstrates an enhancement of up to 8.7, while KFE manifests a maximum improvement of 6.07. Concurrently, we also observed that neither Few-shot nor KFE exhibited a linear improvement with the addition of examples. The performance marginally improved with more than nine examples. In both Few-shot and KFE, the optimal performance is achieved by including nine examples.

\subsubsection*{Effect of Different Instruction Strategies}
To investigate the effectiveness of different instruction strategies $I_{direct}$ and $I_{steps}$, we compared the performance of KFE without and with inference steps. Although prior research has demonstrated generating inference steps significantly improves performance in various reasoning tasks,\cite{cot-first-wei2022} as shown in Table \ref{table: KFE with different settings}, the generation of inference steps reduced performance in the CNMLE. This result also indicated the generation of errors and hallucinations in the reasoning steps and such a limitation that is more serious in knowledge-intensive medical examinations, thus reducing the accuracy. Supplementary Materials 4 and 5 demonstrate case studies of model responses for different instruction and enhancement strategies.

\begin{table}[htbp]\centering
    \caption{Performance of KFE with different settings.}
    \begin{tabular}{lccc}
    \toprule
    \textbf{Setting} & \textbf{Acc-MK(\%)} & \textbf{Acc-CA(\%)} & \textbf{Acc-All(\%)} \\
    \midrule
    \textbf{Instruction strategy} &  &  &  \\
    Direct Instruction $I_{direct}$  & 71.27 & 61.98 & 65.38 \\
    Instruction with inference $I_{steps}$ & 66.30 & 62.62 & 63.97 \\
    \midrule
    \textbf{Knowledge enhancement} &  &  &  \\
    Self-inquiry & 46.96 & 49.84 & 48.79 \\
    Knowledge base & 68.51 & 58.15 & 61.94 \\
    \midrule
    \textbf{Response limitation} &  &  &  \\
    No Limitation & 49.17 & 52.08 & 51.01 \\
    1-token and logit bias & 49.72 & 52.72 & 51.62 \\
    \bottomrule
    \end{tabular}
    \label{table: KFE with different settings}
    \vspace{-0.66cm}
\end{table}

\subsubsection*{Effect of Medical Knowledge Base} 
To investigate the effect of relevant knowledge from the medical knowledge base, we introduce a baseline method \textit{Self-inquiry} adopted in.\cite{llm-knowledge-acl2019, selftalk2020,commonsense2021} Firstly, for each option, we query the GPT with the prompt of \textit{``What is the meaning of \{option\}''} to obtain its meaning; Secondly, we merge all five responses as the internal medical knowledge; Thirdly, we inquire the GPT with the question and generated knowledge. 

As shown in Table \ref{table: KFE with different settings}, with the enhancement of internal knowledge, \textit{Self-inquiry} achieved a score of 48.79 with a 13.15-score reduction. This result suggested that LLM trained on a general domain may lack medical knowledge and \textit{Self-inquiry} does not work in the healthcare domain. Nevertheless, it also demonstrates that the GPT is capable of rapidly digesting and harnessing external knowledge contained in contextual prompts. 


\subsubsection*{Effect of Question Bank Retrieval}
For Few-shot enhancement, we retrieve few-shot examples according to the similarity to the input question (See Method). In this subsection, we compare the performance of pertinent examples with random examples.
Figure \ref{fig: ablation studies}(B) shows a significant reduction in performance for both \textit{Q+O+Correct Ans} and \textit{Q+O+Correct Ans+Generated Inference Detail} when cooperated with random questions, as compared to retrieving related examples from the medical question bank. The former witnessed a decline of 7.89 in the score, whereas the latter experienced a decrease of 6.68. This observation elucidates that the LLM can not only learn the ability of task-solving but also extend their clinical insight derived from few-shot. 


\subsubsection*{Effect of Model Response Length Limitation} 
We set the maximum length of the model response and assign the logit bias of specific characters to constrain the GPT to generate a valid response.\cite{llm-gpt4medical-2023} Specifically, GPT was limited to only generating one token from \textit{\{A, B, C, D, E}\} with equal probability (20\%). As shown in Table \ref{table: KFE with different settings}, this constraint indeed slightly enhanced performance from 51.01 to 51.62 in the Zero-shot setting. However, such limitations would potentially compromise the model's generalizability and impede a fair comparison with others. 


\section*{DISCUSSION}
The rapidly evolving LLMs, bolstered by extensive pre-training and huge model parameters,\cite{llm-encode-knowledge-icml2020, llm-law-2020} encapsulate a wealth of knowledge and exceptional inferential capability.\cite{llm-emergent-2022} Incorporated with digital medicine, LLMs serve as foundational models to underpin the future smart healthcare, spanning clinical decision support,\cite{chatgpt-treatment-ebiomed, decision-chatgpt-jno} online medical consultation,\cite{chatgpt-physician-jno} healthcare surveillance,\cite{chatgpt-advice-lancet} and medical education.\cite{chatgpt-education-2023, chatgpt-education-jama}
This exceptional performance, combined with low barriers to entry, also presents the opportunity to be extended to more lower and middle-income regions, allowing them could effectively benefit from the rapid advancements in digital healthcare and AI.\cite{chatgpt-resource-lancet-2023} 
However, concerns have been raised regarding the predominantly English-centric nature of current advanced models and the prohibitive costs associated with developing LLMs.\cite{llm-nonenglish-book} This disparity could potentially exacerbate inequalities in healthcare services across different countries and languages, highlighting technological and resource imbalances.\cite{chatgpt-resource-lmic}

Our experimental results reveal that general LLMs struggle with non-English clinical question-solving tasks. While ChatGPT and GPT-4 achieved expert-level mastery in USMLE, their efficacy markedly declined when applied to non-English medical examinations.\cite{chatgpt-eye-diagnosis, chatgpt-exam-japan-bias} This discrepancy highlights their inherent limitations in dealing with linguistic diversity, particularly in the healthcare domain.\cite{chatgpt-eye-diagnosis} Therefore, further adaptation are essential for these models to achieve comparable proficiency in non-English contexts.\cite{llm-extend-nonenglish} KFE framework incorporates diverse knowledge resources to construct the Chinese medical context, narrowing the knowledge gap brought by language imbalance. Remarkably, ChatGPT's performance in the CNMLE surged from 51.01 to 70.04, while GPT-4 achieved an expert-level score of 82.59, akin to the performance of Med-PaLM2 in the USMLE.\cite{medpalm2-google-2023} Med-PaLM2 was specifically fine-tuned for English and medical, estimated to cost tens of millions of dollars.\cite{medpalm2-cost-blog} Moreover, this framework enabled a smaller 13B LLM to pass the examination, showing great potential in low-resource settings. 

Despite various strategies have been proposed to enhance LLM's performance,\cite{cot-first-wei2022, incontext-good-2021} their efficacy, particularly in the medical field, remains ambiguous or potentially controversial.\cite{cot-right-2022, cot-false-2022}
Moreover, the requirements of substantial computing resources and the complexity of diverse combinations make it difficult to ascertain the optimal practices in clinical settings. 
This study systematically investigated the effectiveness of different pathways from seven distinct perspectives. Our findings highlight the significant gap in medical knowledge within general LLMs,\cite{clinical-model-knowledge} as evidenced by the comparison of self-inquiry and external knowledge bases. While both knowledge and few-shot enhancement effectively supplement the key clinical insight that had not been adequately encoded during training.\cite{llm-knowledge-wrong} Additionally, whether generating explanations for enriching retrieved examples or providing analysis during question-solving, CoT may lead to a performance decline when compared to direct answer provision. This might be attributed to errors and hallucinations inherent in the generation process of GPTs, emphasizing the knowledge-intensive and rigorous nature of medical tasks. \cite{gpt-pitfalls-2021}
Overall, this study provides practical guidance for enhancing the performance of LLMs in adapting to specific languages or domains. 

Owing to the flexibility of our in-context learning framework, KFE adeptly fuses external medical knowledge from multiple sources. This framework maximally leverages the in-context learning ability of LLMs while obviating the resource-intensive training of LLMs from scratch.\cite{incontext-survey} Therefore, its design supports easy scalability for the incorporation of increasing various types of medical knowledge.\cite{incontext-demo} Furthermore, KFE can be customized to specific clinical specialties or diseases by incorporating corresponding knowledge pools,\cite{llm-rag-disease} thereby enhancing its applicability in diverse clinical contexts.\cite{chatgpt-disease-eye, chatgpt-disease-dentistry}

KFE also supports flexible collaboration with various LLMs. Amidst KFE's support, LLMs ranging from 7 billion to 1.7 trillion parameters demonstrated consistent improvement. This finding not only underscores the pivotal role of knowledge in clinical applications but also highlights the robustness and efficacy of our framework. 
Despite the extensive training and adaptation of BC2-models for the Chinese context, leading to SOTA results in several Chinese language tasks,\cite{baichuan2} its performance still significantly lags behind that of ChatGPT and GPT-4. This gap may be more pronounced in low-resource languages, indicating the necessity of integrating medical knowledge into the advanced LLMs to overcome linguistic barriers in healthcare. 
Although numerous AI-assisted systems of digital medicine have been proposed to reduce healthcare disparities,\cite{dl-expected-reduce-disparities} few have effectively served in clinical settings, especially in resource-limited countries. 
This is attributed to the complexities involved in their deployment and implementation,\cite{medicalai-resource-nmi} coupled with significant performance degradation\cite{model-decline-regions} and potential biases arising from under-represented data.\cite{llm-bias-jno}
These challenges further exacerbate the disparities in healthcare services, preventing global benefits from rapidly evolving AI technologies. \cite{data-imbalance-2023}
In this context, the KFE framework stands out as the most practical solution in resource-limited settings. It facilitates rapid adaptation to evolving LLMs without necessitating additional training, making these models more widely accessible. 
Consequently, the KFE framework presents a promising approach to mitigating global inequalities in digital medicine caused by regional disparities in technological access and linguistic diversity. 

Several potential limitations should be considered for this study. Firstly, the parallel tests were not conducted on more LLMs. This stems from the fact that some large GPT models, such as Med-PaLM(Google), have not yet been available. In addition, the detailed reasoning steps were not further examined from a professional perspective, which could better evaluate the problem-solving capabilities of LLMs. We are collaborating with clinical physicians to further this work, and hope to continue refining it in subsequent studies.

\section*{CONCLUSION}
In conclusion, we investigated the optimal practices for bridging the knowledge gap and enhancing the capabilities of LLMs in non-English medical scenarios. 
By synergizing medical knowledge and few-shot learning through in-context learning, KFE significantly extends the clinical insight of LLMs with varying sizes. It enabled GPT-4 to achieve expert-level performance on the latest CNMLE and also allowed a smaller 13B model to pass the examination, highlighting its potential in resource-limited scenarios. Furthermore, this study sheds light on the effectiveness of diverse pathways for improving model performance. These findings underscore the critical role of integrating medical knowledge into LLMs, thereby fostering their applications in global digital medicine. 
In the era of LLMs, it is imperative to remain cognizant of the potential healthcare disparities arising from technological and linguistic imbalances. Concerted efforts towards enhancing the linguistic versatility and domain-specific expertise of LLMs are essential to improve global digital healthcare, ensuring broader benefits for humanity. 

\newpage
\section*{FUNDING}
This study has no funding support. 

\section*{ACKNOWLEDGEMENTS}
Not applicable.

\section*{AUTHOR CONTRIBUTIONS}
XW, CY, and JY conceived the project. 
JW designed the study, performed the data analysis, and prepared the manuscript.
ZQ, ML, and YZ collected and processed the medical knowledge and medical questions. 
XW, YZ, CY, and JY supervised the study and revised the manuscript. 
All authors contributed to experiments, results interpretation, and final manuscript preparation.

\section*{SUPPLEMENTARY MATERIAL}
Supplementary information includes supplementary figures, tables, and materials (detailed experiment settings, and case studies)

\section*{CONFLICT OF INTEREST STATEMENT}
The authors declare no competing interests. 

\section*{DATA AVAILABILITY}
The code for this study is available in LLM-in-CNMLE [Link of GitHub repository, confidential until paper publication]. Other data that support the findings of this study are available upon reasonable request.

\bibliography{refs}

\appendix


\section*{Supplementary material}

# Large Language Models Leverage External Knowledge to Extend Clinical Insight Beyond Language Boundaries

## Supplementary Information

**Note*:*

*In the actual experiments, all instructions, questions, knowledge, and examples were written in Chinese, without any translation. The English versions in the following images are for easier readability and have been translated using Google Translation.*

# 1. Supplementary Figures

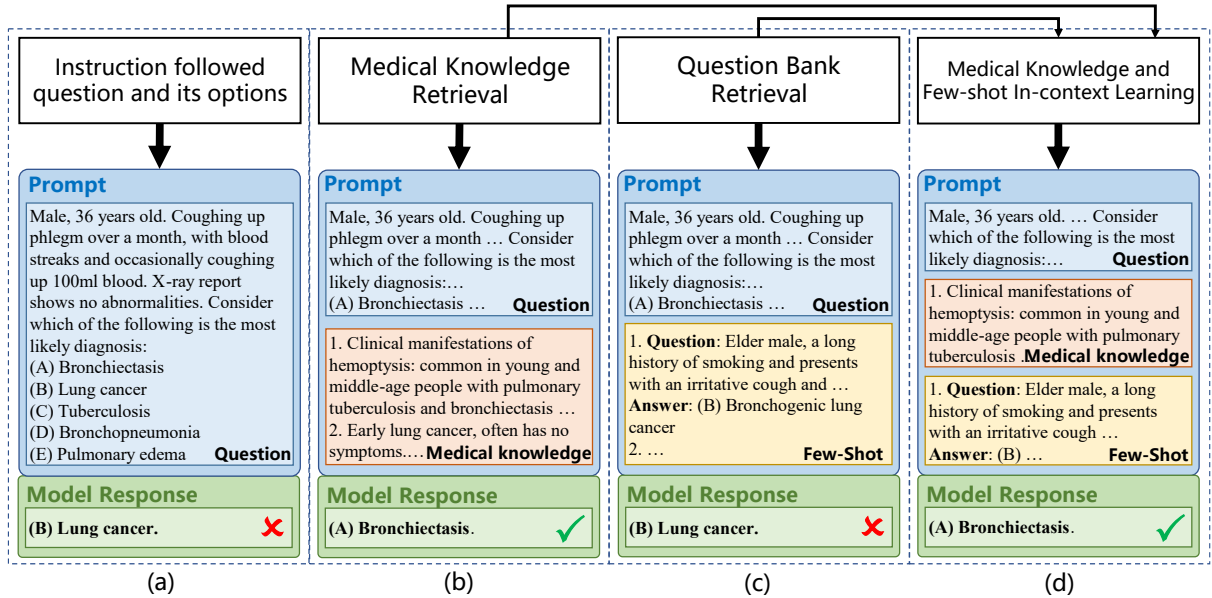

**Supplementary Figure 1. The integrated prompt of different enhancements in KFE.**
(a) list a basic form of prompt that includes the question and options; (b) further includes retrieved related medical knowledge which is in the form of text pieces; (c) includes retrieved pairs of questions and answers as few-shot examples; (d) includes both retrieved knowledge and few-shot examples in prompts.

# 2. Supplementary Tables

**Supplementary Table 1. Model version and access**

| Model | Version | Access |
|---|---|---|
| GPT 3.5 / ChatGPT | gpt-3.5-turbo-0301 | https://platform.openai.com/docs/models/gpt-3-5 |
| GPT4 | gpt-4-0613 | https://platform.openai.com/docs/models/gpt-4-and-gpt-4-turbo |
| Baichuan2-7B | Version 2.0 | https://huggingface.co/baichuan-inc/Baichuan2-7B-Chat |
| Baichuan2-13B | Version 2.0 | https://huggingface.co/baichuan-inc/Baichuan2-13B-Chat |

**Supplementary Table 2. Model parameters**

| Model | Parameter | Version |
|---|---|---|
| GPT-series | max_tokens | 4096 |
| | temperature | 0 |
| | top_p | 1 |
| | frequency_penalty | 0 |
| | presence_penalty | 0 |
| Baichuan-series | model_max_length | 4096 |
| | max_new_tokenss | 512 |
| | do_sample | False |

# 3. Supplementary Materials

**Input of Knowledge Enhancement**

**以下是关于医学知识的单项选择题，请参考给定的相关背景知识，根据问题输出唯一的答案选项**

**背景知识：**

1. 咳嗽咳痰咯血、呼吸困难与水肿 咳嗽咳痰咯血 5.咯血的临床表现 (1) 年龄青壮年咯血多见于肺结核、支气管扩张、二尖瓣狭窄。40岁以上的吸烟者多见于支气管肺癌。

2. 胸心血管外科 肺部疾病 肺癌 临床表现 1.早期肺癌 特别是周围型肺癌往往无任何症状,大多在行胸部X线检查或胸部CT检查时发现。随着肿瘤的进展,出现不同的症状。 血痰常见于中心型肺癌,通常为痰中带血点、血丝或断续地少量咯血,大量咯血少见。

3. 肺结核 1. 可疑症状患者的筛选大约86活动性肺结核患者和95痰涂片阳性肺结核患者有可疑症状主要可疑症状为咳嗽、咳痰持续2周以上和咯血 其次是午后低热、乏力、盗汗、月经不调或闭经有肺结核接触史或肺外结核上述情况应考虑到肺结核病的可能性要进行痰抗酸杆菌和胸部X线检查

4. 细菌性肺炎早期肺脓肿无论在症状和胸部X片表现上与细菌性肺炎很相似，一般细菌性肺炎无大量脓臭痰，胸部X线片示肺叶或段性实变或呈片状淡薄炎症病变边缘模糊不清，没有空腔形成，但当用抗生素治疗高热不退咳嗽、咳痰加剧并咳出大量脓臭痰时应考虑为继发性肺脓肿

5. 临床医学综合 呼吸系统 支气管扩张 诊断与鉴别诊断 4．先天性肺囊肿继发感染时可发生咳嗽咳痰咯血，X线可见多个边界纤细的圆形或椭圆形阴影，周围无浸润，高分辨率CT可明确诊断",

**问题**：男性，36岁。咳痰1月余，痰中有血丝，偶尔咯血100ml，X线检查无异常，考虑诊断为：(A)支气管扩张, (B)肺癌, (C)结核, (D)支气管肺炎, (E)肺水肿

**答案：**

**Model Response of ChatGPT**

A

---

**Input of Knowledge Enhancement**

**Here is a multi-choice question about medical knowledge, please refer to the given relevant background knowledge and examples, and output the unique answer option according to the question.**

**Background knowledge:**

1. Coughing up phlegm and blood, difficulty breathing, and edema. Coughing up phlegm and blood 5. Clinical manifestations of hemoptysis: (1) Hemoptysis is common in young and middle-aged people with pulmonary tuberculosis, bronchiectasis, and mitral valve stenosis. In smokers over 40 years old, it is more common in bronchogenic lung cancer.

2. Thoracic and cardiovascular surgery, lung diseases, lung cancer clinical manifestations: 1. Early lung cancer, especially peripheral lung cancer, often has no symptoms and is mostly found during chest X-ray or chest CT examination. As the tumor progresses, different symptoms appear. Bloody sputum is common in central lung cancer, usually with blood spots, blood streaks in the sputum, or intermittent small amounts of hemoptysis, and large amounts of hemoptysis are rare.

3. Pulmonary tuberculosis: 1. Screening of patients with suspicious symptoms, about 86 active pulmonary tuberculosis patients and 95 sputum smear-positive pulmonary tuberculosis patients have suspicious symptoms. The main suspicious symptoms are cough, phlegm lasting for more than 2 weeks, and hemoptysis, followed by afternoon low fever, fatigue, night sweats, menstrual irregularities or amenorrhea. If there is a history of contact with tuberculosis or extrapulmonary tuberculosis, the possibility of pulmonary tuberculosis should be considered, and sputum acid-fast bacillus and chest X-ray examination should be performed.

4. Bacterial pneumonia: Early lung abscess, both in terms of symptoms and chest X-ray manifestations, is very similar to bacterial pneumonia. General bacterial pneumonia does not have a large amount of foul-smelling pus, and chest X-ray shows lobar or segmental consolidation or patchy thin inflammatory lesions with unclear edges, without cavity formation. However, when high fever persists after antibiotic treatment and cough and sputum worsen and a large amount of purulent sputum is coughed up, secondary lung abscess should be considered.

5. Clinical Medicine Comprehensive, Respiratory System, Bronchiectasis Diagnosis and Differential Diagnosis: 4. Congenital pulmonary cysts can cause cough, phlegm, and hemoptysis when secondary infection occurs. X-rays can show multiple thin-bordered round or oval shadows with no infiltration around them, and high-resolution CT can confirm the diagnosis.

**Question**: Male, 36 years old. Coughing up phlegm over a month, blood-tinged sputum, and occasionally 100ml hemoptysis. X-ray report shows no abnormalities. Consider the following diagnoses: (A) Bronchiectasis (B) Lung cancer (C) Tuberculosis (D) Bronchopneumonia (E) Pulmonary edema

**Answer:**

**Model Response of ChatGPT**

A

**Supplementary materials 1. A case study of the knowledge enhancement.**

Input includes a detailed instruction, related knowledge retrieved from knowledge base, and a testing question.

┌─ Input of Few-shot Enhancement ─────────────────────────────────────────────────────────
以下是关于医学知识的单项选择题，请参考给定的相关例题，根据问题输出唯一的答案选项
参考例题：
(1)问题：老年男性。有长期吸烟史，出现刺激性咳嗽，痰中带血，最可能为: (A)支气管扩张, (B)支气管肺癌, (C)肺炎, (D)肺结核, (E)肺水肿
答案：B
(2)问题：患者，女，58岁，无明显诱因出现反复低热、刺激性咳嗽、咳痰2个月余，痰中可见血丝，体检咳嗽呈高调金属音，抗生素治疗效果欠佳，有多年重度吸烟史。该患者最可能的诊断是: (A)慢性支气管肺炎, (B)支气管肺癌, (C)肺结核, (D)支气管扩张, (E)肺脓肿
答案：B
(3)问题：男性，69岁。持续痰中带血3个月余，既往有慢性支气管炎病史，无支气管扩张、肺结核病史，x线胸片提示右肺中叶占位。该患者咯血的病因是: (A)支气管扩张, (B)肺结核, (C)支气管肺癌, (D)慢性支气管炎, (E)肺脓肿
答案：C
(4)问题：男性，18岁。反复午后发热1个月，体温在37.3～37.8℃，疲乏无力，消瘦。近1周咳嗽，偶尔咯血性痰，夜间盗汗，无胸痛、气短。外院X线检查见右锁骨上斑片状阴影，痰结核菌查阴性。该患者最可能的诊断是: (A)浸润型肺结核, (B)支气管肺癌, (C)支气管扩张合并感染, (D)军团菌肺炎, (E)真菌性肺炎
答案：A
(5)问题：女，22岁，2年来反复痰中带血，间有大口咯血。体格检查无异常体征，X线胸片示左下肺纹理增粗，紊乱，最可能的诊断是: (A)风心病二尖瓣狭窄, (B)慢性支气管炎, (C)支气管扩张症, (D)支气管肺癌, (E)肺结核
答案：C
(6)问题：患者，男性，43岁。3年前患肺结核，抗结核治疗1年，病变吸收好转，痰菌阴转，近3个月来出现间断小量咯血，查痰结核菌阴性，X线胸片双上肺野纤维条索影，患者咯血的原因可能为: (A)肺不张, (B)吸入性肺炎, (C)结核性支气管扩张, (D)支气管扩张, (E)支气管肺癌
答案：C
(7)问题：患者，男性，60岁。吸烟指数大于400，慢性咳嗽咳痰10余年，气促2年，间断血丝痰1月余，近1周出现声嘶，气促，左胸持续性疼痛，食欲下降，消瘦，乏力。测体温37.8℃。本例最可能的诊断是下述哪种疾病: (A)COPD急性加重期, (B)浸润型肺结核, (C)支气管肺癌, (D)支气管哮喘, (E)支气管扩张并咯血
答案：C
(8)问题：女性，18岁，3年来反复咳嗽，大量咳痰，1周前少量咯血，2天前出现高热。首先考虑的诊断是: (A)支气管扩张, (B)支气管内膜结核, (C)慢性阻塞性肺疾病, (D)先天性肺囊肿, (E)早期支气管肺癌
答案：A
(9)问题：女性，19岁，1年来反复痰中带血，偶有咯血。查体无异常，x线胸片示右下肺纹理增粗，紊乱。该患者最可能的诊断为: (A)慢支, (B)copd, (C)哮喘, (D)肺癌, (E)支气管扩张
答案：E

问题：男性，36岁。咳痰1月余，痰中有血丝，偶尔咯血100ml，X线检查无异常，考虑诊断为：(A)支气管扩张, (B)肺癌, (C)结核, (D)支气管肺炎, (E)肺水肿
答案：
├─ Model Response of ChatGPT ─────────────────────────────────────────────────────────────
B
└─────────────────────────────────────────────────────────────────────────────────────────

4

> **Input of Few-shot Enhancement**
>
> **Here is a multi-choices question about medical knowledge, please refer to the given example question and output the unique answer option according to the question.**
> **Examples:**
> (1)Question: Elderly male with a long history of smoking presents with irritating cough and blood-tinged sputum. The most likely diagnosis is:(A) Bronchiectasis, (B) Bronchogenic carcinoma, (C) Pneumonia, (D) Tuberculosis, (E) Pulmonary edema
> Answer: B
> (2)Question: Female, 58 years old, presents with recurrent low-grade fever, irritating cough, and expectoration with streaks of blood for over 2 months, unresponsive to antibiotics, and has a history of heavy smoking for many years. The most likely diagnosis is:( A) Chronic bronchopneumonia, (B) Bronchogenic carcinoma, (C) Tuberculosis, (D) Bronchiectasis, (E) Lung abscess
> Answer: B
> (3)Question: Male, 69 years old, with a history of chronic bronchitis but no history of bronchiectasis or tuberculosis, has had blood in his sputum for over 3 months. X-ray shows an occupying lesion in the right middle lobe of the lung. The cause of his hemoptysis is:( A) Bronchiectasis, (B) Tuberculosis, (C) Bronchogenic carcinoma, (D) Chronic bronchitis, (E) Lung abscess
> Answer: C
> (4)Question: Male, 18 years old, has had recurrent fever in the afternoon for a month, with temperatures between 37.3-37.8°C, fatigue, weight loss, and a recent cough with occasional bloody sputum. He experiences night sweats, but no chest pain or shortness of breath. An X-ray from another hospital shows patchy shadows above the right clavicle, and sputum tests for tuberculosis are negative. The most likely diagnosis is:(A) Infiltrative tuberculosis, (B) Bronchogenic carcinoma, (C) Bronchiectasis with infection, (D) Legionnaires' disease, (E) Fungal pneumonia
> Answer: A
> (5)Question: Female, 22 years old, has had blood in her sputum for 2 years, occasionally coughing up large amounts of blood. Physical examination shows no abnormal signs. X-ray shows increased and disordered lung patterns in the lower left lung. The most likely diagnosis is:(A) Rheumatic heart disease with mitral valve stenosis, (B) Chronic bronchitis, (C) Bronchiectasis, (D) Bronchogenic carcinoma, (E) Tuberculosis
> Answer: C
> (6)Question: Male, 43 years old, was treated for tuberculosis 3 years ago, showed improvement and negative sputum tests after 1 year of anti-tuberculosis treatment. Recently, he has had intermittent small amounts of blood in his sputum. Sputum tests for tuberculosis are negative, and an X-ray shows fibrous streaks in the upper fields of both lungs. The cause of his hemoptysis is likely:(A) Lung atelectasis, (B) Aspiration pneumonia, (C) Tuberculous bronchiectasis, (D) Bronchiectasis, (E) Bronchogenic carcinoma
> Answer: C
> (7)Question: Male, 60 years old, with a smoking index over 400, has had chronic cough and expectoration for over 10 years, shortness of breath for 2 years, and intermittent blood-tinged sputum for over a month. Recently, he has developed hoarseness, shortness of breath, continuous pain in the left chest, loss of appetite, weight loss, and fatigue. His temperature is 37.8 °C. The most likely diagnosis is:(A) Acute exacerbation of COPD, (B) Infiltrative tuberculosis, (C) Bronchogenic carcinoma, (D) Bronchial asthma, (E) Bronchiectasis with hemoptysis
> Answer: C
> (8)Question: Female, 18 years old, has had recurrent cough and large amounts of expectoration for 3 years. She coughed up a small amount of blood a week ago and developed high fever 2 days ago. The first diagnosis to consider is:(A) Bronchiectasis, (B) Endobronchial tuberculosis, (C) Chronic Obstructive Pulmonary Disease, (D) Congenital lung cyst, (E) Early-stage bronchogenic carcinoma
> Answer: A
> (9)Question: Female, 19 years old, has had blood in her sputum for a year and occasionally coughs up blood. Physical examination shows no abnormalities, and an X-ray shows increased and disordered lung patterns in the lower right lung. The most likely diagnosis is:(A) Chronic bronchitis, (B) COPD, (C) Asthma, (D) Lung cancer, (E) Bronchiectasis
> Answer: E
> **Question**: Male, 36 years old. Coughing up phlegm over a month, blood-tinged sputum, and occasionally 100ml hemoptysis. X-ray report shows no abnormalities. Consider the following diagnoses: (A) Bronchiectasis (B) Lung cancer (C) Tuberculosis (D) Bronchopneumonia (E) Pulmonary edema
> **Answer:**
>
> **Model Response of ChatGPT**
> B

## Supplementary materials 2. A case study of the few-shot enhancement.

Input includes a detailed instruction, similar questions retrieved from question bank, and testing question.

**Example**

问题：老年男性。有长期吸烟史，出现刺激性咳嗽，痰中带血，最可能为: (A)支气管扩张, (B)支气管肺癌, (C)肺炎, (D)肺结核, (E)肺水肿

**Question + Options + Correct Answer**

答案：B ✓

**Question + Options + Generated Answer**

答案：根据题目和选项，让我们逐步分析：患者是老年男性，有长期吸烟史，出现刺激性咳嗽，痰中带血。这些症状和病史提示可能是肺结核。肺结核是一种由结核分枝杆菌引起的传染病，常表现为刺激性咳嗽、咳痰，痰中可能带血丝。而支气管扩张通常不会导致痰中带血的症状，支气管肺癌虽然也与吸烟有关，但一般表现为反复咳嗽、咳痰、咳血等症状，但与结核不同，结核通常不会出现高调金属音的体征。肺炎通常伴随发热等症状，与患者的痫史不符。肺水肿通常与心脏问题有关，与吸烟史和痰中带血的症状不太相关。因此，根据题目和选项，最可能的诊断是(D)肺结核。✗

**Question + Options + Generated Correct Answer**

答案：根据题目和选项，让我们逐步分析：老年男性有长期吸烟史，出现刺激性咳嗽，痰中带血，这些症状提示可能是肺部疾病。支气管扩张和肺炎一般不会出现痰中带血的症状，肺水肿也不太可能。肺结核和支气管肺癌都有可能出现刺激性咳嗽和痰中带血的症状，但是肺结核一般不会出现长期吸烟史的情况。因此，最终答案是(B)支气管肺癌。✓

**Question + Options + Correct Answer + Generated Inference Detail**

答案：根据题目中的长期吸烟史和痰中带血的症状，可以初步判断为肺部疾病。支气管扩张和肺炎一般不会出现痰中带血的症状，肺结核和肺水肿也不太可能出现刺激性咳嗽的症状。而支气管肺癌是与吸烟有关的常见疾病，且常伴随着刺激性咳嗽和痰中带血的症状。因此，最终答案是B。✓

**Example**

**Question:** Elderly male with a long history of smoking presents with irritating cough and blood-tinged sputum. The most likely diagnosis is:(A) Bronchiectasis, (B) Bronchogenic carcinoma, (C) Pneumonia, (D) Tuberculosis, (E) Pulmonary edema

**Question + Options + Correct Answer**

**Answer:** B ✓

**Question + Options + Generated Answer**

**Answer:** Based on the question and options, let's analyze step by step: The patient is an elderly male with a long history of smoking, presenting with irritative cough and blood-tinged sputum. These symptoms and medical history suggest the possibility of tuberculosis. Tuberculosis is a contagious disease caused by the Mycobacterium tuberculosis bacterium, commonly manifesting as an irritative cough, expectoration, and possibly streaks of blood in the sputum. While bronchiectasis typically does not lead to blood-tinged sputum, bronchogenic carcinoma, although also related to smoking, generally presents with repeated coughing, expectoration, and hemoptysis, but unlike tuberculosis, tuberculosis usually does not present with a high-pitched metallic percussion sound. Pneumonia usually accompanies symptoms like fever, which does not match the patient's medical history. Pulmonary edema is generally related to heart problems and is not much related to a history of smoking and blood-tinged sputum. Therefore, the most likely diagnosis according to the question and options is (D) Tuberculosis. ✗

**Question + Options + Generated Correct Answer**

**Answer:** Based on the question and options, let's analyze step by step: An elderly male with a long history of smoking presents with irritative cough and blood-tinged sputum. These symptoms suggest a potential pulmonary disease. Bronchiectasis and pneumonia generally do not present with blood-tinged sputum, and pulmonary edema is also unlikely. Both tuberculosis and bronchogenic carcinoma may present with irritative cough and blood-tinged sputum, but tuberculosis generally does not occur with a long history of smoking. Therefore, the final answer is (B) Bronchogenic carcinoma. ✓

**Question + Options + Correct Answer + Generated Inference Detail**

**Answer:** Based on the long history of smoking and blood-tinged sputum symptoms in the question, it can be preliminarily judged as a pulmonary disease. Bronchiectasis and pneumonia generally do not present with blood-tinged sputum, and both tuberculosis and pulmonary edema are unlikely to present with irritative cough symptoms. Bronchogenic carcinoma, a common disease related to smoking, often accompanies symptoms like irritative cough and blood-tinged sputum. Therefore, the final answer is B. ✓

**Supplementary materials 3. A case study of different strategies for few-shot enhancement.**

**Question**

问题：男性，36岁。咳痰1月余, 痰中有血丝,偶尔咯血100ml, X线检查无异常, 考虑诊断为:(A)支气管扩张, (B)肺癌, (C)结核, (D)支气管肺炎, (E)肺水肿
答案：

**Zero-shot**

B ✗

**Knowledge Enhancement**

A ✓

**Few-shot Enhancement**

B ✗

**Knowledge and Few-shot enhancement**

A ✓

**Question**

Question: Male, 36 years old. Coughing up phlegm over a month, blood-tinged sputum, and occasionally 100ml hemoptysis. X-ray report shows no abnormalities. Consider the following diagnoses: (A) Bronchiectasis (B) Lung cancer (C) Tuberculosis (D) Bronchopneumonia (E) Pulmonary edema
Answer:

**Zero-shot**

B ✗

**Knowledge Enhancement**

A ✓

**Few-shot Enhancement**

B ✗

**Knowledge and Few-shot enhancement**

A ✓

**Supplementary materials 4. A case study of the predicted answer by GPT3.5.**

**Question**

问题：男性，36岁。咳痰1月余, 痰中有血丝,偶尔咯血100ml, X线检查无异常, 考虑诊断为:(A)支气管扩张, (B)肺癌, (C)结核, (D)支气管肺炎, (E)肺水肿
**答案:**

**Zero-shot**

根据题目和选项，让我们逐步分析：根据症状描述，咳痰1月余，痰中有血丝，偶尔咯血100ml，考虑到肺部疾病。X线检查无异常，但并不能完全排除肺部疾病的可能性。*支气管扩张一般不会出现咯血症状*，肺水肿也不符合症状描述。支气管肺炎一般不会持续1个月以上，因此排除。结核和肺癌都有可能导致咳痰、咯血等症状，但结核一般不会出现大量咯血，而肺癌则有可能。因此，最终答案是**(B)肺癌**。 ✗

**Knowledge Enhancement**

根据题目和背景知识，让我们逐步分析：*根据年龄和咯血量，该患者可能患有支气管扩张或肺癌。但X线检查无异常，排除了肺癌的可能性*。支气管肺炎和肺水肿的症状与咯血不符，因此也可以排除。结核患者通常会有咳嗽持续2周以上的症状，但该患者只有咳痰1月余，因此结核的可能性较小。根据支气管扩张的临床表现，该患者的症状与支气管扩张相符，因此最终答案是**(A)支气管扩张**。 ✓

**Few-shot Enhancement**

根据题目，该患者咳痰1月余，痰中有血丝，偶尔咯血100ml，X线检查无异常。因此，可以排除肺水肿的可能性。*支气管扩张和肺炎一般不会出现痰中带血的症状*，肺结核可能出现痰中带血，但一般不会出现偶尔咯血100ml的情况。而肺癌是导致痰中带血和咯血的常见病因之一，且X线检查可能无异常。因此，最终答案是**B肺癌**。 ✗

**Knowledge and Few-shot enhancement**

根据背景知识和例题，让我们逐步分析：该患者出现咳痰1月余，痰中有血丝，偶尔咯血100ml，X线检查无异常。*根据背景知识可知，咯血量小量咯血为<100ml/d，因此该患者咯血量属于小量*。年龄为36岁，*不符合肺癌和肺水肿的典型发病年龄*。结核和支气管肺炎也可能引起咳痰和痰中带血的症状，但X线检查无异常，*不符合结核和支气管肺炎的典型表现*。而*支气管扩张也可能引起咳痰和痰中带血的症状，且常见于年轻人*，因此最可能的诊断是支气管扩张，选项A正确。因此，最终答案是**A**。 ✓

**Question**

**Question:** Male, 36 years old. Coughing up phlegm over a month, blood-tinged sputum, and occasionally 100ml hemoptysis. X-ray report shows no abnormalities.
Consider the following diagnoses: (A) Bronchiectasis (B) Lung cancer (C) Tuberculosis (D) Bronchopneumonia (E) Pulmonary edema
**Answer:**

**Zero-shot**

Based on the question and options, let us analyze step by step: According to the symptom description, the patient has had a cough with phlegm for over a month, with blood streaks in the phlegm and occasionally coughing up 100ml of blood. This indicates a lung disease. X-ray examination shows no abnormalities, but it cannot completely rule out the possibility of lung disease. *Bronchiectasis generally does not present with hemoptysis symptoms*, and pulmonary edema also does not match the symptom description. Bronchopneumonia generally does not last for more than a month, so it is ruled out. Both tuberculosis and lung cancer may cause symptoms such as cough with phlegm and hemoptysis, but tuberculosis generally does not present with a large amount of hemoptysis, while lung cancer may. Therefore, the final answer is **(B) lung cancer**. ✗

**Knowledge Enhancement**

Based on the question and background knowledge, let us analyze step by step: *According to the age and amount of hemoptysis, the patient may have bronchiectasis or lung cancer*. However, *the X-ray examination shows no abnormalities, ruling out the possibility of lung cancer.* The symptoms of bronchopneumonia and pulmonary edema do not match hemoptysis, so they can also be ruled out. Tuberculosis patients usually have a cough lasting more than 2 weeks, but this patient has had a cough with phlegm for over a month, so the likelihood of tuberculosis is low. According to the clinical manifestations of bronchiectasis, the patient's symptoms are consistent with bronchiectasis, so the final answer is **(A) bronchiectasis**. ✓

**Few-shot Enhancement**

According to the question, the patient has had a cough with phlegm for over a month, with blood streaks in the phlegm and occasionally coughing up 100ml of blood. The X-ray examination shows no abnormalities. Therefore, the possibility of pulmonary edema can be ruled out. *Bronchiectasis and pneumonia generally do not present with bloody phlegm symptoms.* Tuberculosis may present with bloody phlegm, but it usually does not involve occasionally coughing up 100ml of blood. Lung cancer is one of the common causes of bloody phlegm and hemoptysis, and the X-ray examination may show no abnormalities. Therefore, the final answer is **B lung cancer**. ✗

**Knowledge and Few-shot enhancement**

Based on background knowledge and examples, let us analyze step by step: The patient presents with a cough with phlegm for over a month, blood streaks in the phlegm, and occasionally coughing up 100ml of blood. The X-ray examination shows no abnormalities. *According to the background knowledge, a small amount of hemoptysis is defined as <100ml/d, so the patient's hemoptysis is considered a small amount.* The patient is 36 years old, which *does not match the typical onset age for lung cancer and pulmonary edema*. Tuberculosis and bronchopneumonia may also cause symptoms of cough with phlegm and bloody phlegm, but the X-ray examination shows no abnormalities, which *does not match the typical presentation of tuberculosis and bronchopneumonia*. *Bronchiectasis may also cause symptoms of cough with phlegm and bloody phlegm and is more common in young people*. Therefore, the most likely diagnosis is bronchiectasis, and option A is correct. Thus, the final answer is **A**. ✓

**Supplementary materials 5. A case study of the generated inferences by GPT3.5.**

The *red italic* represents the incorrect information, while the *green italic* denotes the key rationales underlying the correct answer. The *blue italic* and *brown italic* indicate the rationales derived from knowledge enhancement or few-shot enhancement, respectively.



\end{document}